\documentclass[twocolumn]{article}
\usepackage{caption}
\usepackage{multirow}
\usepackage{esvect}
\usepackage{bigfoot}
\usepackage{etoolbox}

\makeatletter
\patchcmd{\maketitle}{\@fnsymbol}{\@alph}{}{}  
\makeatother
\usepackage{floatrow}
\newfloatcommand{capbtabbox}{table}[][\FBwidth]
\usepackage{tabularx}

\usepackage{blindtext}
\usepackage{geometry}                
\usepackage{amsmath}

\DeclareMathOperator*{\argmaxA}{arg\,max} 

\usepackage{hyperref}

\usepackage{verbatim}

\hypersetup{
    colorlinks=true,
    linkcolor=blue,
    filecolor=magenta,
    urlcolor=blue,
}
\usepackage{multicol}
\usepackage{mathptmx}       
\usepackage{helvet}         
\usepackage{courier}        
\usepackage{type1cm}        

\usepackage{enumitem}

\usepackage{makeidx}         
\usepackage{graphicx}        
\usepackage{multicol}        
\usepackage[bottom]{footmisc}

\makeindex             

\title{The Training of Neuromodels for Machine Comprehension of Text. Brain2Text Algorithm.
}

\makeatletter
\renewcommand\@date{{%
  \vspace{-\baselineskip}%
  \large\centering
  \begin{tabular}{@{}c@{}}
    A. Artemov \textsuperscript{1} \\
  \end{tabular}%
  \quad  \quad
  \begin{tabular}{@{}c@{}}
    A.Sergeev \textsuperscript{1} \\
  \end{tabular}
  \quad  \quad
  \begin{tabular}{@{}c@{}}
    I. Khasenevich \textsuperscript{1} \\
  \end{tabular}
  \quad  \quad
  \begin{tabular}{@{}c@{}}
    A. Yuzhakov \textsuperscript{2} \\
  \end{tabular}
  \quad  \quad
  \begin{tabular}{@{}c@{}}
    M. Chugunov\textsuperscript{2} \\
  \end{tabular}

  \bigskip

\begin{tabular}{@{}c@{}}
    \textsuperscript{1} Cognitive Systems Company \footnote{With the support of the Foundation for Assistance to Small Innovative Enterprises, Russia}  \\
    \normalsize science@cogsys.company 
\end{tabular}%
  \quad                           \quad
\begin{tabular}{@{}c@{}}
    \textsuperscript{2} Promobot Company \\
    \normalsize info@promo-bot.ru
\end{tabular}%

  \bigskip

}}
\makeatother

\begin{document}



\maketitle
{\bf Keywords:} machine learning, text recognition, neural networks, one-shot learning, large neural networks, classification.

\abstract{Nowadays, the Internet represents a vast informational space, growing exponentially and the problem of search  for relevant data  becomes essential as never before. The algorithm proposed in the article allows to perform natural language queries on content of  the document and get comprehensive meaningful answers. The problem is partially solved for English as SQuAD contains  enough data to learn on, but there is no such dataset in Russian, so the methods used by scientists now are not applicable to Russian. Brain2 framework allows to cope with the problem - it stands out for its ability to be applied on small datasets and does not require impressive computing power. The algorithm is  illustrated on Sberbank of Russia Strategy's text and assumes the use of a neuromodel consisting of 65 mln synapses. The trained model is able to construct word-by-word answers to questions based on  a given text. The existing limitations are its current inability to identify synonyms, pronoun relations and allegories. Nevertheless,  the results of conducted experiments showed  high capacity and  generalisation ability of  the suggested approach.}

\section{Introduction}
Although, natural language processing  is one of the most rapidly developing fields in computer science, reading comprehension  and questions answering still remain areas where human outperforms any model aimed to comprehensively understand text . The main difficulty for a long time was the lack of a dataset of an appropriate quality  and size to train on. An attempt to resolve  the issue was made in [Rajpurkar el al., 2016] \cite{latexcompanion}, They  designed  a text corpus based on Wikipedia's articles,  which  contains a total of 23 215 paragraphs.
Authors developed a logistic regression model trained on correct answers to questions based on the content of the given paragraphs, which is able to answer  any question from any other text source. An important specification of the model is that it is able to match nouns, related pronouns and also it distinguishes synonyms. The data for the training set was obtained by crowd-working - people were paid for constructing questions and answers  on given paragraphs.  The features related to matching words, bigrams, roots and other lexical specifications of question and answer were used. F1-score of 0,51 was obtained while for  humans this measure equals  0,86. Also, [Wang, Jiang, 2016] \cite{latex} presented a model architecture  based on a match-LSTM  whose F1 score is 71\%  on the test dataset.
Another hot issue  in engineering  of linguistic models  is the number of minimal basis elements, by which the text will be divided. For example, in  Word2Vec\footnote{Mikolov et al. Vector Representation of Words. https://www.tensorflow.org/tutorials/word2vec} models like SkipGram and CBOW it is a  5-word 'window' and 4 adjoint words which account for the context. On this problem  the following hypothesis was proposed by authors:
\textbf{\textit{Two words are enough to determine the  unique meaning of the third word}}

 It can be interpreted by the Euclidian geometry in the following way: "If we span any word in the decomposition in terms of basis of  two given words, then  its semantic in this basis is represented as a  line, which intersects (0,0) and coordinates of the word with that meaning".
 Based on this intuition, several algorithms were developed,  mutually constituting  a complex sequence-to-sequence model.

\section{Algorithm description}

The process of getting an answer consists of 4 stages.  7 models are used during the whole iteration:
\begin{enumerate}
  \item  The first algorithm determines \textbf{parts of speech (POS)} which should  be included in the answer.\\ Words from the question are used as input, e.g. such as "When"/ "Who" / "How" / "What happened" / "Which one" and returns "Numerical"/ "Noun" / "Adverb" / "Noun" / "Adjective"  respectively.
  \item \textbf{Minimal linguistic semantic unit (MLSU).} \\ This  one gets in the input question's tokens and required POS, and returns the ID of MLSU.
  \item \textbf{Verb determining algorithm.} \\ It takes MLSU ID  and returns the verb that is best suitable for the answer  \item \textbf{MLSU tokens determination.}  \\ The model gets MLSU id and  returns the set of tokens to  be used in the answer.
  \item \textbf{Next token determination.}  \\ To the input   the current and 2 previous words are given,  and the model returns the token (or set of tokens) which should go next.
  \item \textbf{Previous token determination.} \\  Inversed version of previous algorithm.
  \item  \textbf{Token-to-word model.} \\ The inputs are: adjacent words, token, token's POS, MLSU's verb. It returns the next word that will be used in the answer.
\end{enumerate}
  \begin{figure}
  \centering

  \includegraphics[scale=.25]{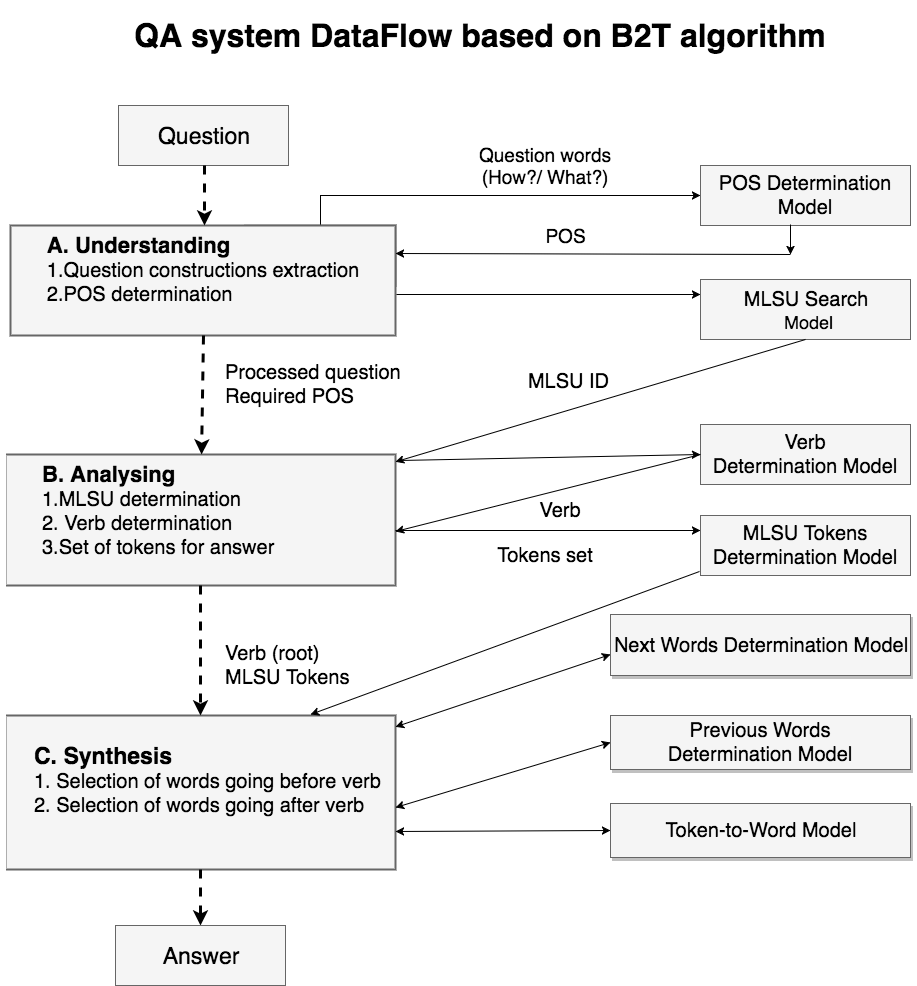}
  \end{figure}

  At the first stage, the question is being preprocessed. It is separated into two parts - informative and interrogative. From the informative part MLSUs are extracted (they have the  following structure : "verb (or nouns)  + START + contextual tokens  + END").

Further, the POS of the searched item (X) is being determined using the interrogative part.

In the next stage, using MSLUs and X's POS  the algorithm determines the ID of the content which is represented as a set of tokens.

And finally, the sentence is being synthesised in the following way: the token of the nex word is chosen using two previous words. After that, we use the token and previous words to determine the final form of the word. The process starts form the verb as it is stored as a word, not a token. 

Example: John listens  to classical music every day while his sisters listen  to emo.

There are two MLSU:

1. MLSU ID1= [ listens; Context ID1=\{John; classical; music; every; day, to, START, END\}]

2. MLSU ID2=[ listen; Context ID2=\{ while, he, sister, emo, to, START, END\}] \newline%

Each one consists of a verb and set of tokens, and together they constitute a semantic unit. 

The algorithm iterates over all MLSUs and after each iteration the  chosen word is removed from the set until it gets START or END. \newline%

For example, for the first MLSU we have tokens {John; classical; music; every; day,to,START,END\} \newline%
\begin{enumerate}
\item Algorithm starts iterating  from ” none” + ”listen” and gets ”to”. ("to" is a token)
\item The pair  "listen"  and "to" gives the final form of 'to' - "to".
\item Then, using "listen" and "to" we obtain "emo".
\item The same procedure is repeated until  we get END.
\end{enumerate}

So we have  reduced our set  {while, he, sister, ,START} for choice of the previous token and its word-shaped form.  Then we will go from right to left:
\begin{enumerate}
\item 'To'  and 'listen' give  'sister'.
\item From  'listen' and 'sister' the model returns 'sisters'  and so on \footnote{For Russian lexemes were used instead of tokens due to linguistic characteristics
.}.

\end{enumerate}


%

\section{Model Training}
\subsection{Informational Neurobayesian Approach}

Weights in the model are calculated using our special approach named Informational Neurobayesian Approach (INA), in which every weight represents  quantity of information of the the object's feature $i$ which activates given neuron $j$,(Pointwise Mutual Information modification).
$$ I_{ij} = \psi_{ij} \  \log{\frac{P_{ij}}{P_j}} + b,   \ \  b = I_0 - bias  $$
where $ \psi$ - coefficient of emergence of the system for class (layer)\footnote{[Lutsenko E.V. 2002]} \cite{luc} $j$ and feature $i$.
$ \psi_{ij} =\frac{log_2 (2^{W_i}-1)}{\log N_j},$ where $N$ - number of possible  conditions of the system (outputs), $W_i$ - number of features, figuring in the decision process; bias is a parameter for activation of  class. Emergence coefficient represents the  information, obtained from synthesis of several classes, which was not available before.

Thorough description of the approach can be found in [Artemov et al., 2017] \cite{art}.

\begin{figure}
\centering
\caption{Summation Process Architecture}

\includegraphics[scale=.3]{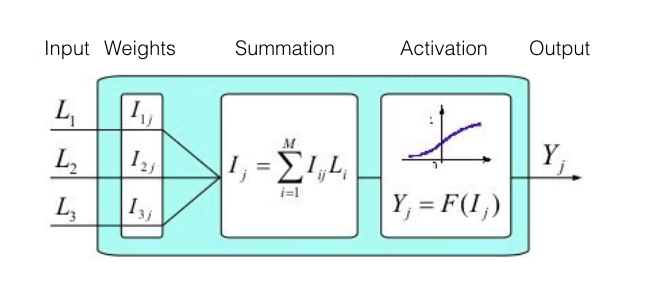}
\end{figure}
That neural network can be illustrated by the Table 1.
\begin{table}
\centering
\caption{Table representation of the knowledge \newline%
neuromodel}
\label{my-label}
 \begin{tabular}{|c|c|c|l|c|c|c|}
\hline
\multicolumn{2}{|l|}{\multirow{2}{*}{}} & \multicolumn{5}{c|}{Classes}   \\ \cline{3-7}
\multicolumn{2}{|l|}{}                  & 1    & ... & j    & ... & W    \\ \hline
\multirow{5}{*}{Features}     & 1       & $I_{11}$ &     & $I_{1j}$ &     &$ I_{1W}$ \\ \cline{2-7}
                              & ...     &      &     &      &     &      \\ \cline{2-7}
                              & $i$       & $I_{i1}$ &     & $I_{ij}$ &     & $I_{iW}$ \\ \cline{2-7}
                              & ...     &      &     &      &     &      \\ \cline{2-7}
                              & $M$       & $I_{M1}$ &     & $I_{Mj}$ &     & $I_{MW}$ \\ \hline
\multicolumn{2}{|l|}{Number of    }                       & $N_1$ &     & $N_j$ &     & $N_W$ \\
\multicolumn{2}{|l|}{ objects of class $j$}                       &  &     &  &     & \\  \hline

 \end{tabular}
\end{table}

In the scheme above the $L_i$ is $i$-the object's feature. $\vv{L}\{\alpha_1, ...,\alpha_w\} $-- all its features.

\begin{equation}
    \vv{L}_i  \ = \
    \begin{cases}
      1, & \text{if} \  \alpha <\epsilon  \\
      0, & \text{otherwise}
    \end{cases}
\end{equation}

For every object with a given set of features the neural network  chooses over all classes the one with the maximal informational criteria, represented as an activated sum of all features.
$$  j* = \argmaxA_j (\vv{I_{ij}}, \vv{L_i}) $$ $$
Y=F(I_{j*}) $$
The Y is a chosen class for the given features.

\subsection{Training}

The problem can formulated as to  train the model on natural text and design the algorithm of  lexeme determination, which  together will be used to answer natural language questions. So, at first, the algorithm  chooses the words to use in the answer and at the next stage composes   coherent sentence as an answer .

There are 5 types of models designed for problem's solution. The first one determines which part of speech is  not present in the question but should be included in the final answer. The second is seeking for words. Two models are designed to build sequence of words in the sentence  and the last one to get a token from the word.

The training process in the first model  goes in the following way:

\textit{\{Why it is light during daytime?  -- Sun shines \}}.  Full answer:  \textit{\{It is light at daytime, because the sun shines\} } $\rightarrow$ X  is Action (A) \newline%
There were  7 types of answers predetermined for the model:
\begin{enumerate}
\item  O -- object ( nouns, pronouns);
\item OD-- object's description ( adjective, numerical, participle, gerund);
\item  S -- subject (noun, pronoun);
\item SD -- subject's description ( adjective, numerical, participle, gerund);
\item A -- action ( verb);
\item  AD - actions's description ( pronoun);
\item OT - other.
\end{enumerate}

The unknown word represents action, the X is a verb.
The input data is preprocessed so that  interrogative and contextual parts are extracted from it. This is  what allows to train the model to distinguish the type of word to search for.
The data for the first stage of the training is  shown in Table 2.

\begin{table}
\centering
\caption{A fragment of the knowledge neuromodel for choosing a part of speech}
\label{my-label}
 \begin{tabular}{|c|c|c|c|}
\hline
\multirow{ 2}{*}{Features}&Intervals of           &\multicolumn{2}{c|}{Intervals of Classes}      \\ \cline{3-4}
   &Features & Adjective & Noun      \\ \hline
 &\multicolumn{1}{l|}{Bias}               & 0         & 0         \\
                      & \multicolumn{1}{l|}{which}           & 0,534   & -0,239   \\
                      & \multicolumn{1}{l|}{by whom}              & 0,040   & 0,191    \\
Question & \multicolumn{1}{l|}{when}            & 0         & 0,092    \\
construction& \multicolumn{1}{l|}{to whom}             & -0,186  & 0,244   \\
                      & \multicolumn{1}{l|}{who}              & 0         & 0,261   \\
                      & \multicolumn{1}{l|}{what}          & 0,117   & 0,164   \\
                      & \multicolumn{1}{l|}{what is}         & 0,274   & -0,032  \\
                      & \multicolumn{1}{l|}{whose}              & 0         & 0,216     \\ \hline
  & \multicolumn{1}{l|}{Bias}             & 0         & 0     \\
Lexical part  & \multicolumn{1}{l|}{any}      & 0,064   & 0,101   \\
of POS& \multicolumn{1}{l|}{adverb}        & 0,025   & -0,050 \\
                      & \multicolumn{1}{l|}{adjective}    & 0,084   & -0,004  \\ \hline
 \end{tabular}
 \newline%
 \newpage
\end{table}
\begin{table}
 \begin{tabular}{|c|c|c|c|}
\hline
\multirow{ 2}{*}{Features}&Intervals of           &\multicolumn{2}{c|}{Intervals of Classes}      \\ \cline{3-4}
   &Features& Any                   & Adverb       \\ \hline
 &\multicolumn{1}{l|}{Bias}            & 0                     & 0               \\
                      & \multicolumn{1}{l|}{which}          & 0,649               & 0,113    \\
                      & \multicolumn{1}{l|}{by whom}        & 0,555               & 0         \\
Question & \multicolumn{1}{l|}{when}           & 0                     & 0,643    \\
construction& \multicolumn{1}{l|}{to whom}        & 0                     & 0           \\
                      & \multicolumn{1}{l|}{who}            & 0,658               & 0           \\
                      & \multicolumn{1}{l|}{what}           & 0                     & 0           \\
                      & \multicolumn{1}{l|}{what is}        & 0                     & -0,061   \\
                      & \multicolumn{1}{l|}{whose}          & 0,894               & 0            \\ \hline
  & \multicolumn{1}{l|}{Bias}           & 0                     & 0             \\
Lexical part  & \multicolumn{1}{l|}{any}            & 0,111                & -0,361   \\
of POS& \multicolumn{1}{l|}{adverb}         & -0,108              & -0,052  \\
                      & \multicolumn{1}{l|}{adjective}      & 0,026               & -0,144   \\ \hline
 \end{tabular}
\end{table}

Now using the sentence "The sun shines at morning and men go to work" as an example of  the framework will be demonstrated in Table 3.

\begin{table*}[h]

\centering
\caption{Data for training}
\label{my-label}
 \begin{tabular}{l|l|l|l}
\multirow{2}{*}{Model} & \multirow{2}{*}{Training data}
{\begin{tabular}[c]{@{}l@{}} \end{tabular}}                                                        & \multicolumn{2}{c}{ \ \ \ \ \ \ \ \ \ \ \ \  Structured   \ data}                                                                                                                                                                              \\
                       &                                                                                                                                          &  \multicolumn{1}{c|}{Features}                                                                                                        & \multicolumn{1}{c}{Classes}                                                                   \\ \hline
0     & \begin{tabular}[c]{@{}l@{}}Q: Why it is light at morning?\\ A. The sun shines.\\ \\ Q: Where do men go?\\ A: Men go to work\end{tabular} & \begin{tabular}[c]{@{}l@{}}Question + content tokens: \\  \\"Why" +  \{adverb, noun\}\\ "Where" + \{verb, men\}\end{tabular} & \begin{tabular}[c]{@{}l@{}}Unknown POS: Verb\\ Unknown POS : Noun\end{tabular}       \\ \hline
1     & \begin{tabular}[c]{@{}l@{}}The sun shines at morning.\\ Men go to work\end{tabular}                                                      & \begin{tabular}[c]{@{}l@{}}Shines (morning, sun)\\ Go (to, men, work)\end{tabular}                                        & \begin{tabular}[c]{@{}l@{}}Content ID: \\ MLSU ID\_X\\ MLSU ID\_\{X+1\}\end{tabular} \\ \hline
2     & \begin{tabular}[c]{@{}l@{}}Initial sequence of\\  three tokens:\\ \{Man, go,  to\}\end{tabular}                                       & \begin{tabular}[c]{@{}l@{}}Tokens and their POS: \\ \textit{man\_noun,  go\_verb}.\end{tabular}                                    & \begin{tabular}[c]{@{}l@{}}Next token: \\ \textit{to}\end{tabular}                           \\ \hline
3     & \begin{tabular}[c]{@{}l@{}}Initial sequence of\\  three tokens\\ \textit{\{The, sun, shines\}}\end{tabular}                                       & \begin{tabular}[c]{@{}l@{}}Tokens and their POS : \\ \textit{sun\_noun, shines\_verb}.\end{tabular}                                & \begin{tabular}[c]{@{}l@{}}Previous token:\\ \textit{the}\end{tabular}             \\ \hline
4     & \begin{tabular}[c]{@{}l@{}}Initial sequence of \\ words' pairs\\ \textit{Sun shines, men work}\end{tabular}                                       & \begin{tabular}[c]{@{}l@{}}Token from MSLU\_ID
 \\ and its POS:\\ \textit{ sun\_noun, to\_any}\end{tabular}                           & \begin{tabular}[c]{@{}l@{}}Next token:\\ \textit{shines, work}\end{tabular}                   \\ \hline
5     & \begin{tabular}[c]{@{}l@{}}Initial sequence\\  of words' pairs: \\ \textit{The sun, men go }\end{tabular}                                      & \begin{tabular}[c]{@{}l@{}}Token from MSLU\_ID and it's POS: \\ \textit{sun\_noun, to\_any}\end{tabular}                           & \begin{tabular}[c]{@{}l@{}}Previous token:\\ \textit{the, men}\end{tabular}
 \end{tabular}
\end{table*}
Authors will be glad to provide access to the data corpus for the model's training at reader's request.

\section{Experiment Results}
Sberbank's strategy text was used for model training. The text is in Russian and consists of 656 sentences, total volume of 8200 words, 2048 tokens and 514 verbs. Brain2 framework was used to  design the network, total number of connections accounted for over 65 mln. The model parameters are shown in Table 4. \newline%
The original  document is available at \href{http://www.sberbank.ru/common/_en/img/uploaded/files/SberbankDevelopmentStrategyFor2014-2018_en.pdf}{Sberbank's website.}

\footnote{\tiny{\verb| sberbank.ru/common/_en/img/uploaded/files/SberbankDevelopmentStrategyFor2014-2018_en.pdf|}}  }
The algorithm's execution is demonstrated in the   web-interface, accessible by the URL: \url{sb.brain2.online}.
\begin{figure}
\caption{Framework Demo}

\includegraphics[scale=.2]{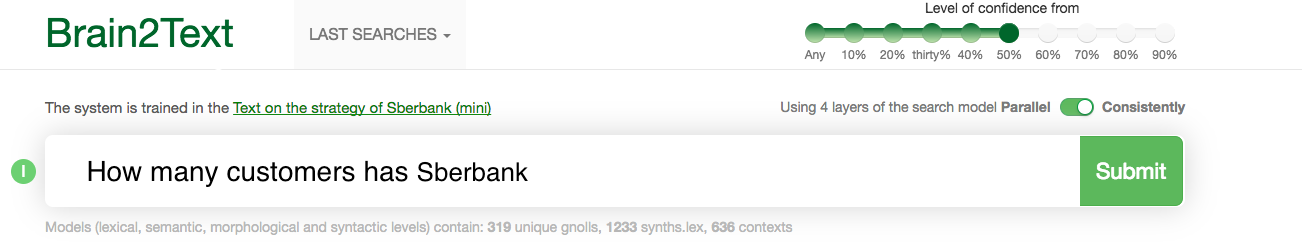}

\includegraphics[scale=.2]{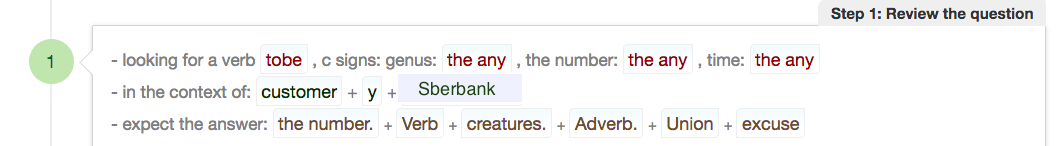}

\includegraphics[scale=.2]{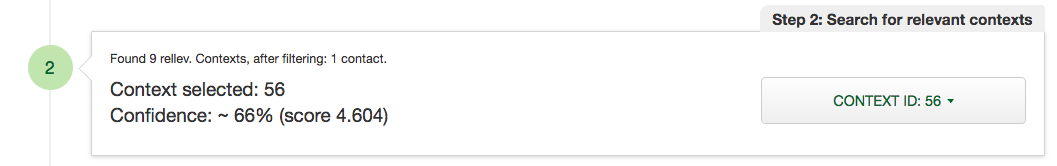}

\includegraphics[scale=.2]{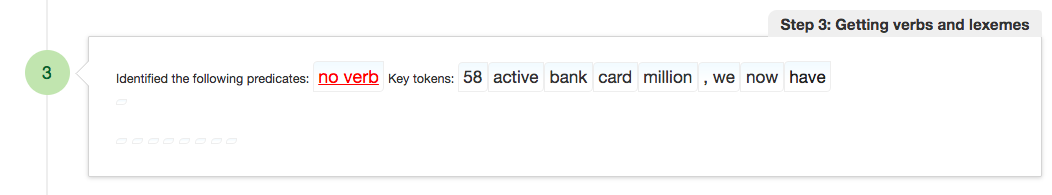}

\includegraphics[scale=.2]{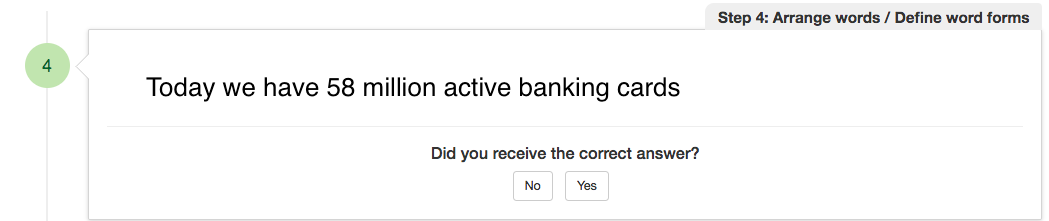}

\end{figure}
\newpage
\begin{table}
\centering
\caption{Model's parameters}
\label{my-label}
 \begin{tabular}{c|c|c|c|l|c|c}
Model & Size & Number   & Number       & Number of    & F1-measure     &Precision /      \\
           &         &    of   classes &  of features &  connections &              & Recall \\ \hline
\multicolumn{1}{l|}{1. Question Processing}&\multicolumn{1}{r|}{10 Kb}&\multicolumn{1}{r|}{12}&\multicolumn{1}{r|}{133}&\multicolumn{1}{r|}{1 596}&\multicolumn{1}{r|}{0,24} & \multicolumn{1}{r}{0,34 /0,1} \\  \hline
\multicolumn{1}{l|}{2. Word Search}&\multicolumn{1}{r|}{1 Mb}&\multicolumn{1}{r|}{636}&\multicolumn{1}{r|}{1 577}&\multicolumn{1}{r|}{1 002 972}&\multicolumn{1}{r|}{0,56} &\multicolumn{1}{r}{0,62 / 0,5}\\  \hline
\multicolumn{1}{l|}{3. Text Composing} &\multicolumn{1}{r|}{23 Mb}&\multicolumn{1}{r|}{1488}&\multicolumn{1}{r|}{8 031 }&\multicolumn{1}{r|}{11 950 128 }& \multicolumn{1}{r|}{0,750 } & \multicolumn{1}{r}{0,87 / 0,6} \\
 \multicolumn{1}{l}{Right-to-Left} &      &                   &                    &               &        &\multicolumn{1}{r}{} \\ \hline
  \multicolumn{1}{l|}{4. Text Composing} &\multicolumn{1}{r|}{23 Mb}&\multicolumn{1}{r|}{1464}&\multicolumn{1}{r|}{8 069}&\multicolumn{1}{r|}{11 813 016}&\multicolumn{1}{r|}{0,771} &\multicolumn{1}{r}{0,91 / 0,6} \\
 \multicolumn{1}{l}{Left-to-Right}&      &  &                    &             &          & \\ \hline

 \multicolumn{1}{l|}{5. Next word} &\multicolumn{1}{r|}{77 Mb}&\multicolumn{1}{r|}{2656}&\multicolumn{1}{r|}{15 118}&\multicolumn{1}{r|}{40 153 408}&\multicolumn{1}{r|}{0,98} & \multicolumn{1}{r}{ 0,99/ 0,9} \\
 \multicolumn{1}{l}{determination}&      &                   &                    &                       & &

 \end{tabular}
\end{table}
\subsection{Tests}

Three groups of questions were designed to train on: 1) Questions based on the content of the text. The correctness of answers was checked against them.
2) Questions on irrelevant topics. 3) Meaningless questions (for tuning Type II error).
Also, more than 6000 questions were designed automatically, dividing 3 groups in the same way.

The interface was designed to illustrate the process the system is going through.

Every obtained answer was compared with the original one, and in case of match 1 point was assigned,  half points were assigned if an answer was classified as alternative and 0 otherwise.
The confidence was calculated as a fraction  from the mean information on a feature.
Integral estimate is a dot product of points and their confidences. Consecutive training approach gives better results.

\begin{table}[b]
\centering

\caption{Expert questions-based testing results}
\label{my-label}
 \begin{tabular}{l|l|l}
Parameter & Parallel & Consecutive  \\
\hline
Questions asked&30 &30\\ \hline
Correct answers&   18 &   24\\ \hline
Correct answers  &8.9 &19.2\\
(integral estimate)         &      &         \\ \hline
                            &      &         \\ \hline
Type I Error &   48\% &   23\%\\ \hline
Type II Error &  $<   1\%$ &   $<   1\%$\\ \hline
Type I Error  &   81\% &   41\%\\
(Integral measure)       &      &        \\
\hline
Type II Error  &   $<   1\%$ &  $<  1\%$\\
(Integral measure)     &      &

 \end{tabular}
\end{table}

\begin{table}
\centering

\caption{Technical questions-based testing results}
\label{my-label}
 \begin{tabular}{l|l|l}
Parameter & Parallel & Consecutive  \\
\hline
Questions asked&6000 &6000\\ \hline
F - measure &  0.83681  &   0.88736\\ \hline
Precision  & 0.85206 & 0.8624\\
Recall &0.82208&0.91382\\ \hline
 \end{tabular}
\end{table}
\section{Conclusion}
The results of the experiments confirmed the validity of the two-words hypothesis. Presented natural language processing model is able to answer  questions  with precision rate of 0.822-0.914. The model has is not able yet to distinguish synonyms, pronouns and allegories. Nevertheless, given the current restrictions it shows quite promising results. In the future, it is planned  to develop the algorithm  to recognize synonyms, pronouns etc, and  also make it available in English.


\begin{thebibliography}{1}

\bibitem{latexcompanion}
Pranav~Rajpurkar, Jian~Zhang, Konstantin~Lopyrev and Percy~Liang.
\newblock {\em SQuAD: 100,000+ Questions for Machine Comprehension of Text.}
\newblock Arxiv, 2016.

\bibitem{latex}
Shuohang~Wang and Jing~Jiang.
\newblock {\em Machine Comprehension Using Match-LSTM and Answer Pointer.}
\newblock Arxiv, 2016.

\bibitem{luc}
Lutsenko~E.V.
\newblock {\em Conceptual principles of the system (emergent) information theory and its application for the cognitive modelling of the active objects (entities).}
\newblock Computer society, 2002.

\bibitem{art}
A.~Artemov, E.~Lutsenko, E.~Ayunts and I.~Bolokhov.
\newblock {\em Informational Neurobayesian Approach to Neural Networks Training. Opportunities and Prospects.}
\newblock Arxiv, 2017.

\end{thebibliography}
\end{document}